\documentclass{article}

\PassOptionsToPackage{}{natbib}
\usepackage[dblblindworkshop,final]{neurips_2026}
\workshoptitle{Med-Reasoner}

\usepackage[utf8]{inputenc}
\usepackage[T1]{fontenc}
\usepackage{graphicx}
\usepackage{amsmath,amssymb,amsfonts}
\usepackage{booktabs}
\usepackage{multirow}
\usepackage[table]{xcolor}
\usepackage{listings}
\usepackage{subcaption}
\usepackage{xspace}

\usepackage{siunitx}
\usepackage{array}
\usepackage[normalem]{ulem}
\usepackage{adjustbox}

\definecolor{bestgray}{gray}{0.90}

\newcommand{\best}[1]{{\cellcolor{bestgray}\bfseries #1}}

\usepackage{soul}
\usepackage{colortbl}
\usepackage{changepage,threeparttable}
\usepackage{enumerate}
\usepackage{enumitem}

\usepackage{hyperref}
\usepackage{url}

\lstdefinelanguage{smtlib}{
  morekeywords={declare-const,declare-fun,assert,check-sat,get-model,
                set-logic,push,pop,Bool,Int,Real,and,or,not,=,ite,
                forall,exists,ALL},
  sensitive=true,
  morecomment=[l]{;},
  morestring=[b]",
  alsoletter={-},
}
\title{Toward Guarantees for Clinical Reasoning in Vision Language Models via Formal Verification}

\author{%
  Vikash Singh\thanks{Equal contribution.},
  Debargha Ganguly\footnotemark[1],
  Haotian Yu, Chengwei Zhou,\\
  \textbf{Prerna Singh, Brandon Lee, Vipin Chaudhary, Gourav Datta}\\
  Case Western Reserve University\\
  \texttt{\{vikash,debargha,haotian.yu,chengwei.zhou,prerna,}\\
  \texttt{brandon.lee,vipin,gourav.datta\}@case.edu}
}

\begin{document}

\maketitle
\begin{abstract}
Vision-language models (VLMs) show promise in drafting radiology reports, yet they frequently suffer from logical inconsistencies, generating diagnostic impressions unsupported by their own perceptual findings or missing logically entailed conclusions. Standard lexical metrics heavily penalize clinical paraphrasing and fail to capture these deductive failures in reference-free settings. Toward guarantees for clinical reasoning, we introduce a neurosymbolic verification framework that deterministically audits the internal consistency of VLM-generated reports. Our pipeline autoformalizes free-text radiographic findings into structured propositional evidence, utilizing an SMT solver (Z3) and a clinical knowledge base to verify whether each diagnostic claim is mathematically entailed, hallucinated, or omitted. Evaluating seven VLMs across five chest X-ray benchmarks, our verifier exposes distinct reasoning failure modes, such as conservative observation and stochastic hallucination, that remain invisible to traditional metrics. On labeled datasets, enforcing solver-backed entailment acts as a rigorous post-hoc guarantee, systematically eliminating unsupported hallucinations to significantly increase diagnostic soundness and precision in generative clinical assistants.
\end{abstract}

\section{Introduction}
\label{sec:intro}
\begin{figure}[htbp]
    \centering
    \includegraphics[width=1\linewidth]{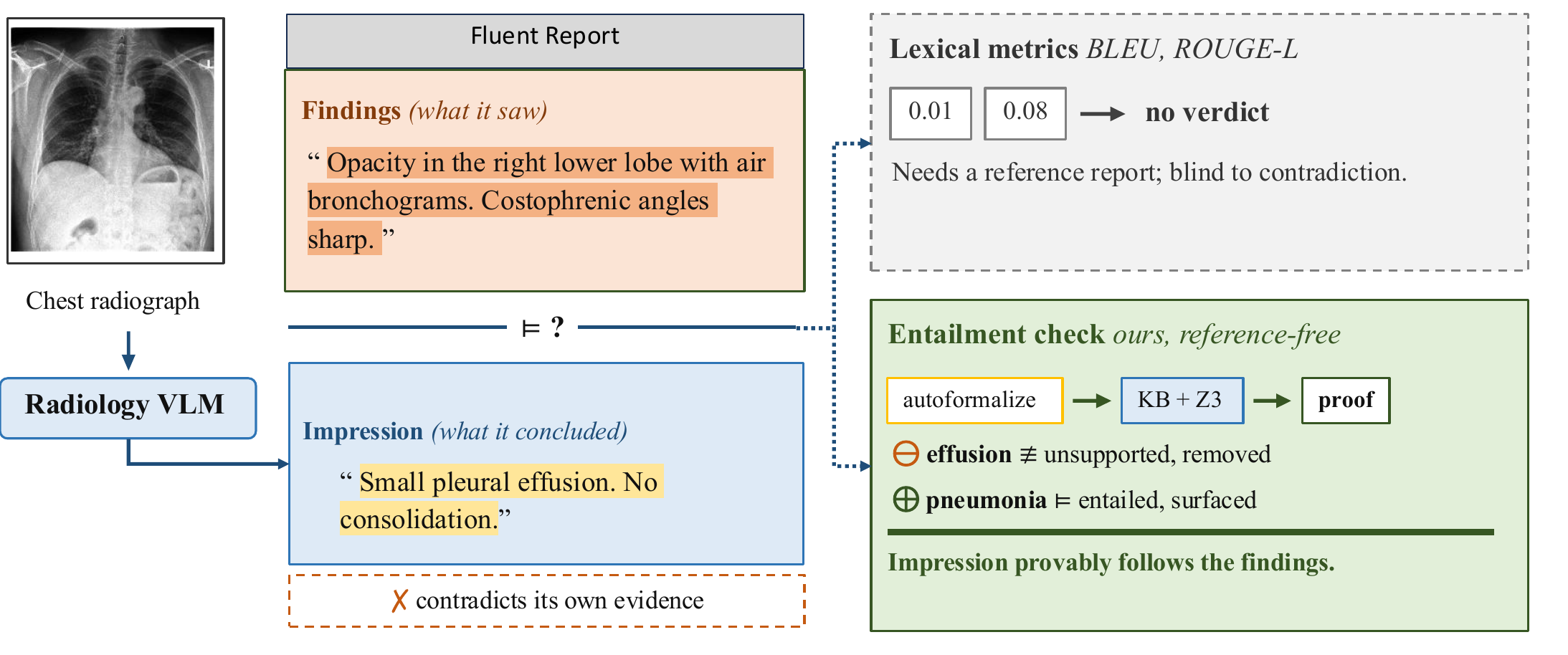}
\caption{Autoregressive VLMs generate fluent diagnostic impression that logically contradicts their own perceptual findings. Our neurosymbolic framework extracts structured evidence from the text and utilizes Z3 to deterministically prove whether a diagnostic claim is logically entailed or hellucinated where standard lexical metrics are blind to these deductive failures,}
    \label{fig:motivation}
\end{figure}
Unprecedented imaging volumes have accelerated the adoption of generative vision-language models (VLMs) as cognitive assistants in radiology~\citep{brady2020artificial, moor2023foundation}. By drafting preliminary reports from visual context, models like MedGemma and LLaVA-Med aim to reduce clinician fatigue and turnaround times~\citep{tu2024towards}. However, deploying these models in safety-critical clinical environments exposes a fundamental vulnerability: the absence of formal guarantees. In medicine, statistical plausibility is an insufficient proxy for correctness; practitioners must trust that generated drafts are deductively valid. Today, this trust is unfounded, as VLMs function as probabilistic text generators optimized for fluency rather than verifiable clinical reasoning~\citep{bender2021dangers}.

The core technical barrier to building verifiable VLMs lies in their autoregressive objective function~\citep{brown2020language}. A model generates a report by maximizing the likelihood of the next token, modeled as $P(w_t | w_{<t}, I)$. While this effectively captures correlations between visual features and clinical phrasing~\citep{radford2021learning}, it fundamentally precludes intrinsic mechanisms to guarantee logical entailment~\citep{marcus2019rebooting}. Consequently, a VLM may correctly perceive visual evidence in the ``Findings'' section (e.g., ``blunted costophrenic angle'') yet fail to deduce the mathematically forced conclusion in the ``Impression'' (e.g., ``pleural effusion''). Conversely, it may hallucinate unsupported diagnoses based on training priors~\citep{ji2023survey}. This stochasticity creates a dangerous ``illusion of reasoning''~\citep{valmeekam2022large}, entirely negating the technology's assistive value when clinicians must expend effort debugging fluent but fundamentally contradictory logic~\citep{goddard2012automation}.

Compounding this issue, current evaluation paradigms are entirely empirical and fail to capture deductive validity. Standard NLP metrics (BLEU~\citep{papineni2002bleu}, ROUGE~\citep{lin2004rouge}), entity-aware scores (F1-CheXbert~\citep{smit2020chexbert}) and ood methods~\citep{ganguly2026trust,11433375,10.1145/3785462.3815821,ganguly2025forte} operate strictly on reference similarity against a ground-truth report. Unlike traditional software validation, which leverages formal methods to prove system properties, these metrics cannot assess the internal logical consistency of a generated report in isolation and it leaves them blind to explicit contradictions (Figure ~\ref{fig:motivation}). In real-world clinical workflows where ground truth is intrinsically unavailable, these metrics offer no safety guarantees. A system that generates a fluent but self-contradictory report poses a severe risk of automation bias~\citep{parasuraman2010complacency}, underscoring the urgent need to transition from empirical string-matching to formal verification.

Toward providing formal guarantees for clinical reasoning, we propose a novel neurosymbolic verification framework that imposes formally verifiable constraints on the output of VLMs~\citep{hitzler2022neural, hamilton2024neuro}. Drawing on formal verification principles, we treat the internal logical consistency of a drafted radiology report as a formal satisfiability (SAT) problem~\citep{biere2021preprocessing}. As illustrated in Figure ~\ref{fig:motivation}, Our pipeline explicitly decouples visual perception, handled probabilistically by the VLM, from clinical reasoning, which is audited deterministically by a symbolic solver~\citep{ganguly2024proof, singh2026verge} and causal systems~\cite{singh2026causalguardconformalinferencegraph}. Our contributions are:

\begin{enumerate}[nosep,leftmargin=1.3em]
    \item We introduce a reference-free neurosymbolic framework that bridges probabilistic text generation and deterministic logic. By mapping free-text findings into SMT constraints via a clinical ontology, we enable the automated, runtime verification of diagnostic logic \emph{independently of ground-truth reference texts}.
    \item Through the formal auditing of seven VLMs across five chest X-ray benchmarks, we expose distinct deductive failure modes, ranging from conservative observation to stochastic hallucination, that empirical, reference-based metrics are mathematically incapable of detecting.
    \item We demonstrate that applying an SMT solver as a post-hoc deductive safeguard on labeled datasets provably eliminates unsupported hallucinations. This enforcement of mathematical entailment yields targeted, quantifiable guarantees for diagnostic soundness and precision in generative clinical assistants.
\end{enumerate}

\section{Related Work}
\paragraph{Factual Consistency in Clinical Report Generation.}
Generative vision-language models ~\citep{bordes2024introductionvisionlanguagemodeling} for automated radiology reporting frequently produce fluent text that lacks clinical grounding. To suppress factual errors, existing frameworks rely on empirical interventions, including semantic entity rewards~\cite{delbrouck2022improvingfactualcorrectnessradiology}, multi-level reinforcement learning~\cite{zhao2026hallucinationmitigatingmedicalreport}, and uncertainty quantification via semantic consistency across sampled generations~\cite{wang2025semanticconsistencybaseduncertaintyquantification}. While these methods improve statistical alignment and suppress unsupported claims on average, they remain probabilistic heuristics that evaluate drafts against reference reports. Consequently, they cannot provide deductive guarantees or assess whether an impression logically follows from observed findings in reference-free clinical deployment.

\paragraph{Neurosymbolic Verification and Solver-Aided Reasoning.}
To overcome the deductive vulnerabilities of autoregressive models, recent neurosymbolic methods pair neural decoders with automated theorem provers and SMT solvers. In general reasoning, frameworks such as VeriCoT~\cite{davidovich2026stochbenchdomainspecificbenchmarkstochastic,feng2026vericot} and NSF-CoT~\cite{nsfcot} validate natural language chains of thought by formalizing reasoning steps into first-order logic and checking entailment with SMT solvers. Similarly,~\cite{sistla2025verifiedcodereasoningllms} verify neural explanations against formal program semantics. Beyond static checking, interactive agents synthesize checkable formal artifacts via SMT-LIB scratchpads~\cite{anonymous2026sequentprover}, while autoformalization pipelines leverage generative reward models~\cite{anonymous2026beyond,min2026theorylevelautoformalizationisolatedstatements} and consistency distillation~\cite{hou2026stratifiedconsistencydistillationnatural} to bridge informal language and symbolic specifications. Closely related formal engines deploy solvers for runtime verification and deductive refinement~\cite{bayless2025neurosymbolic,ganguly2025grammars,singh2026verge,ganguly2024proof}. We extend this paradigm to multimodal clinical workflows, using SMT solving to deterministically audit deductive entailment between perceptual findings and diagnostic impressions.
\section{Methodology}
\label{sec:method}
\begin{figure}[htbp]
    \centering
    \includegraphics[width=1\linewidth]{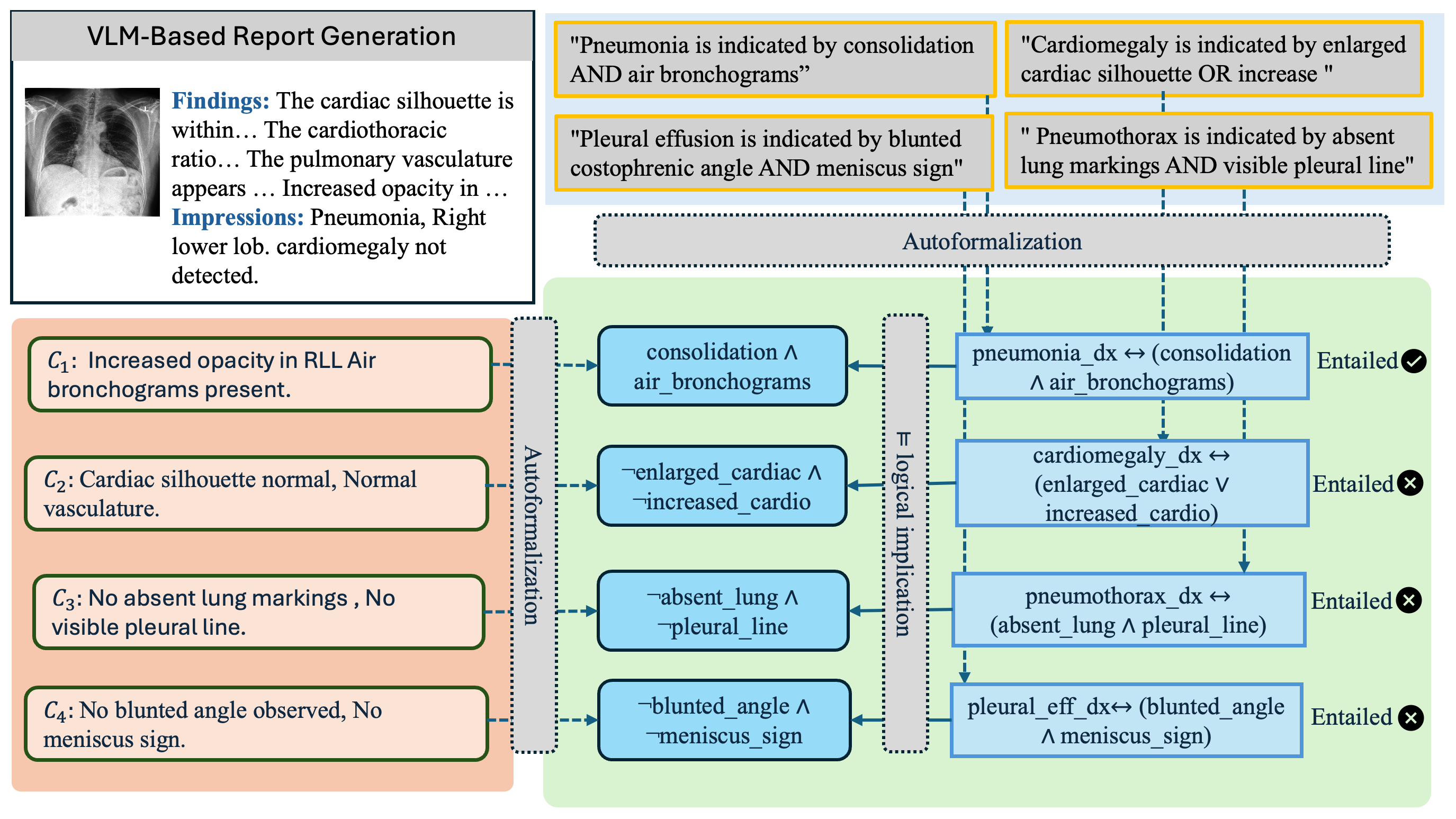}
\caption{Neurosymbolic verification pipeline. A VLM generates Findings and Impression text, which are autoformalized into structured evidence and checked against a clinical knowledge base using Z3 to verify clinical reasoning.}
    \label{fig:methodology}
\end{figure}
Given an image $I$ and a VLM $\mathcal{M}_\theta$, the model generates a report $R = (R_F, R_I)$ comprising a \emph{Findings} section $R_F$ and an \emph{Impression} section $R_I$. Our objective is to verify whether the diagnostic claims in $R_I$ are logically entailed by the perceptual evidence asserted in $R_F$ under a fixed clinical knowledge base, effectively isolating visual perception from deductive reasoning to provide formal guarantees of internal consistency.

\subsection{Ontological Grounding and Autoformalization}
To transition from free text to a computable domain, we define a lightweight formal ontology $\mathcal{O}=\langle\mathcal{F}, \mathcal{D}, \mathcal{K}\rangle$. $\mathcal{F}=\{f_1,\ldots,f_N\}$ is a finite set of atomic observational predicates (e.g., \texttt{costophrenic\_blunting}), $\mathcal{D}=\{d_1,\ldots,d_M\}$ is a set of diagnostic predicates (e.g., \texttt{pleural\_effusion}), and $\mathcal{K}$ represents the clinical knowledge base, modeled as a set of propositional formulas defining sufficient conditions and consistency constraints for each diagnosis (e.g., $d \Rightarrow \neg d'$). To construct $\mathcal{K}$, we employ a large language model (GPT-5.2) to autoformalize established clinical guidelines into candidate propositional rules, which are subsequently audited and refined by a team of clinicians to guarantee ontological correctness and exhaustiveness.

Because deterministic verification over natural language is intractable, we introduce an autoformalization function $\mathcal{T}$ that maps $R_F$ to a structured evidence assignment. We implement $\mathcal{T}$ for evaluation purposes under a closed-world assumption (CWA), mirroring the clinical reporting standard where significant findings not mentioned are implicitly deemed absent. $\mathcal{T}$ yields a binary state vector $V \in \{1,0\}^{|\mathcal{F}|}$, where for each $f_i \in \mathcal{F}$:
\[
V_i=\begin{cases}1 & \text{if } f_i \text{ is explicitly affirmed in } R_F,\\0 & \text{otherwise (explicitly negated, unmentioned, or underspecified).}\end{cases}
\]
We implement $\mathcal{T}$ via a strictly constrained LLM (GPT-OSS-20B, temperature $0.0$) enforcing a schema-aligned JSON output to ensure deterministic structural parsing. We analogously extract the asserted diagnoses $D(R_I) \subseteq \mathcal{D}$ from the \emph{Impression} section via strict schema-constrained string matching.

\subsection{Diagnostic Entailment via Satisfiability}
We frame report verification as a formal satisfiability (SAT) problem. Let $\Phi_V$ denote the propositional context induced by the autoformalized findings vector $V$, constructed as a conjunction of all asserted literals:
\[
\Phi_V\equiv\bigwedge_{\{i \mid V_i=1\}} f_i \land \bigwedge_{\{i \mid V_i=0\}} \neg f_i
\]

For any claimed diagnosis $d \in D(R_I)$, we test if $d$ is a logical consequence of the evidence under the knowledge base ($\Phi_V \land \mathcal{K} \models d$). We compile $\mathcal{K}$ into SMT constraints and use the Z3 solver as a deterministic decision procedure to check the satisfiability of the negated conclusion:
\[
\textsc{IsSat}(\Phi_V\land\mathcal{K}\land\neg d)
\]
Because the SMT solver is sound and complete, this formulation provides a mathematical proof of entailment, directly inducing a reference-free, runtime-deployable taxonomy of reasoning errors strictly based on internal consistency:
\begin{enumerate}
    \item \textbf{Supported (Entailed):} $d \in D(R_I)$ and the check is \textsc{Unsat}. Every valid model of the stated Findings forces the diagnosis $d$.
    \item \textbf{Unsupported (Hallucinated):} $d \in D(R_I)$ but the check is \textsc{Sat}. The diagnosis is asserted, but there exists a valid state where the Findings hold and $d$ does not.
    \item \textbf{Missed (Omitted):} $d \in \mathcal{D} \setminus D(R_I)$ and the check is \textsc{Unsat}. The diagnosis is logically forced by the evidence but absent from the Impression.
    \item \textbf{Correctly Excluded:} $d \in \mathcal{D} \setminus D(R_I)$ and the check is \textsc{Sat}. The diagnosis is neither forced by the evidence nor claimed.
\end{enumerate}

\noindent \textbf{Verifier Reliability.} Deductive guarantees follow a formal assume-guarantee paradigm, conditioned on the VLM's visual grounding of $R_F$ and the translational fidelity of $\mathcal{T}$. In this assistive workflow, the radiologist validates perceptual assertions, while autoformalization uncertainty is quantified and surfaced via grammar entropy~\citep{ganguly2025grammars} during clinical review.

\section{Experiments and Analysis}
\label{sec:experiments}

\subsection{Datasets and VLMs}
We evaluate whether neurosymbolic verification can accurately characterize reasoning quality and improve diagnostic performance in radiology reports. 

\noindent\textbf{Models.} Our experiments span multiple model families and medical adaptation regimes, benchmarking three general-purpose VLMs (Qwen3-VL-8B~\citep{bai2025qwen3vltechnicalreport}, Llava-Mistral-7B, Llava-Vicuna-7B ~\citep{liu2023llava,li2023llavamed,liu2023improvedllava}) and four domain-adapted medical VLMs (Med Gemma-4B, MedGemma-27B~\citep{medgemma2024}, Lingshu-7B, Lingshu-32B~\citep{lasateam2025lingshugeneralistfoundationmodel}).

\noindent\textbf{Datasets.}We test on standard chest radiograph corpora divided into two evaluation regimes. First we look at paired-report datasets, but without ground truth classification diagnosis. specifically MIMIC-CXR~\citep{johnson2019mimic}, Indiana-IU, and CheXpert-Plus~\citep{chambon2024chexpert}. These are used to compute baseline lexical overlap against ground-truth human reports. Next, we look at datasets with ground truth classification labels, CheXpert~\citep{irvin2019chexpert}, NIH-CXR\cite{8099852} to quantify diagnostic performance using structured ground-truth labels. For each image, models are prompted to generate a report containing distinct \emph{Findings} ($R_F$) and \emph{Impression} ($R_I$) sections. We apply the autoformalizer $\mathcal{T}$ to extract a structured findings assignment $V$ and an impression diagnosis set $D(R_I)$. All verification is performed using the Z3 solver with the fixed clinical knowledge base $\mathcal{K}$.

\subsection{Baseline: Lexical Overlap}
We first evaluate report quality using standard $n$-gram overlap metrics (BLEU and ROUGE-L) on the paired-report datasets. As shown in Table~\ref{tab:nl_metrics_overall}, overlap scores are near-zero across all model families and scales. Clinically equivalent phrasing (e.g., ``blunted costophrenic angle'' vs.\ ``trace pleural effusion'') is heavily penalized by these metrics. The severe drop between Findings ROUGE-L and Impression ROUGE-L further illustrates that lexical similarity is an inadequate proxy for evaluating deductive clinical reasoning.

\begin{table}[htbp]
\centering
\begin{tabular}{
p{3.1cm}
S[table-format=1.4]
S[table-format=1.4]
S[table-format=1.4]
S[table-format=1.4]
S[table-format=1.4]
S[table-format=1.4]
}
\toprule
& \multicolumn{2}{c}{Findings}
& \multicolumn{2}{c}{Impression}
& \multicolumn{2}{c}{Full Report} \\
\cmidrule(lr){2-3}
\cmidrule(lr){4-5}
\cmidrule(lr){6-7}
Model
& {BLEU} & {ROUGE-L}
& {BLEU} & {ROUGE-L}
& {BLEU} & {ROUGE-L} \\
\midrule
Lingshu-7B
& 0.0235 & \best{0.1875} & 0.0114 & 0.0787 & \best{0.0205} & \best{0.1719} \\
Lingshu-32B
& \best{0.0237} & 0.1809 & \best{0.0155} & \best{0.0860} & 0.0196 & 0.1667 \\
Llava-Mistral-7B
& 0.0040 & 0.0703 & 0.0029 & 0.0229 & 0.0039 & 0.0689 \\
Llava-Vicuna-7B
& 0.0051 & 0.0834 & 0.0028 & 0.0368 & 0.0046 & 0.0794 \\
MedGemma-4B
& 0.0138 & 0.1295 & 0.0081 & 0.0731 & 0.0127 & 0.1236 \\
MedGemma-27B
& 0.0110 & 0.1266 & 0.0034 & 0.0379 & 0.0108 & 0.1258 \\
Qwen3-VL-8B
& 0.0032 & 0.0670 & 0.0018 & 0.0256 & 0.0020 & 0.0437 \\
\bottomrule
\end{tabular}
\vspace{1em}
\caption{BLEU and ROUGE-L overlap between generated and ground-truth reports across Findings, Impression, and Full Report sections, showing that lexical metrics poorly reflect clinical reasoning quality.}
\label{tab:nl_metrics_overall}
\end{table}

\subsection{Reference-Free Auditing of Internal Consistency}
To evaluate internal logic without relying on external ground truth, we apply our neurosymbolic verifier to test entailment between the generated evidence ($\Phi_V$) and the claimed diagnoses ($D(R_I)$). We formally define two novel reference-free metrics to quantify a model's deductive reliability. The knowledge base for each dataset was curated separately and audited by a clinician; it covers the diagnosis labels required by that dataset and includes the major diagnostic rules/conditions based on expert opinion.

Let $E_V$ represent the set of all diagnoses logically entailed by the observed findings under the knowledge base, defined as $E_V = \{ d \in \mathcal{D} \mid \Phi_V \land \mathcal{K} \models d \}$. We define \textbf{Soundness} ($S$) as the proportion of generated impression claims that are logically supported by the findings, and \textbf{Completeness} ($C$) as the proportion of logically entailed diagnoses that the model successfully verbalizes:

\[
S = \frac{| E_V \cap D(R_I) |}{| D(R_I) |} ; C = \frac{| E_V \cap D(R_I) |}{| E_V |}
\]

Inspired by the concept of soundness in formal logic, we target $S \ge \textbf{0.99}$~\citep{bayless2025neurosymbolic} for safety-critical clinical deployments, ensuring that incorrect or unsupported validity claims are exceedingly rare.

\begin{table}[htbp]
\centering
\begin{adjustbox}{max width=\linewidth}
\begin{tabular}{
p{3.4cm}
S[table-format=4.0]
S[table-format=1.4]
S[table-format=1.4]
S[table-format=1.4]
S[table-format=1.4]
S[table-format=1.4]
S[table-format=1.4]
}
\toprule
Model & {N} & {Precision} & {Recall} & {F1} & {Soundness} & {Completeness} & {Specificity} \\
\midrule
\texttt{Lingshu-7B} &
4056 & 0.5340 & 0.3588 & 0.4292 & 0.9562 & 0.9103 & 0.9491 \\

\texttt{Lingshu-32B} &
4062 & 0.6183 & 0.4293 & 0.5068 & 0.9636 & 0.9217 & 0.9579 \\

\texttt{Llava-Mistral-7B} &
4059 & 0.5323 & 0.2409 & 0.3317 & 0.9702 & 0.8930 & 0.9653 \\

\texttt{Llava-Vicuna-7B} &
4061 & 0.4749 & 0.1997 & 0.2812 & 0.9555 & 0.8387 & 0.9442 \\

\texttt{MedGemma-4B} &
4050 & 0.6759 & 0.5432 & 0.6023 & 0.9652 & 0.9389 & 0.9598 \\

\texttt{MedGemma-27B} &
4062 & 0.7850 & \best{0.7113} & \best{0.7463} & 0.9811 & \best{0.9721} & 0.9791 \\

\texttt{Qwen3-VL-8B} &
3787 & \best{0.9436} & 0.4279 & 0.5888 & \best{0.9943} & 0.8735 & \best{0.9927} \\
\bottomrule
\end{tabular}
\end{adjustbox}
\vspace{1em}
\caption{Reference-free evaluation (aggregated as mean across datasets) of deductive reliability using neurosymbolic entailment. Precision, recall, F1, soundness, completeness, and specificity quantify how well Impression diagnoses are logically supported by generated Findings. }
\label{tab:your_label_here}
\end{table}
Applying this logical audit exposes three distinct VLM failure modes that remain invisible to standard lexical metrics. We first observe clinically consistent models (e.g., MedGemma-27B). These models exhibit balanced, high soundness and completeness. Claims are reliably supported by findings, and logically forced conclusions are consistently verbalized in the impression. Next, there are conservative observer models (e.g., Qwen3-VL-8B). These models optimize for soundness at the expense of completeness. The model rarely hallucinates an unsupported diagnosis, but frequently fails to deduce and report diagnoses that are mathematically forced by its own observed findings. Finally, the highest amount of stochasticity comes in models like Llava-Vicuna-7B. They are characterized by low precision and low completeness, these models treat the reporting task as statistical text generation rather than grounded reasoning, frequently hallucinating impression statements entirely unsupported by prior evidence. We additionally performed manual spot-checking of 100 text-to-SMT translations by formal verification experts and found no translation errors in this sampled set. As is standard in formal verification, our guarantees are conditional (precondition/postcondition style): correctness holds relative to the assumed correctness of the text-to-symbolic translation and the curated knowledge base, rather than as an unconditional clinical guarantee.
\begin{figure}
    \centering
    \includegraphics[width=\linewidth]{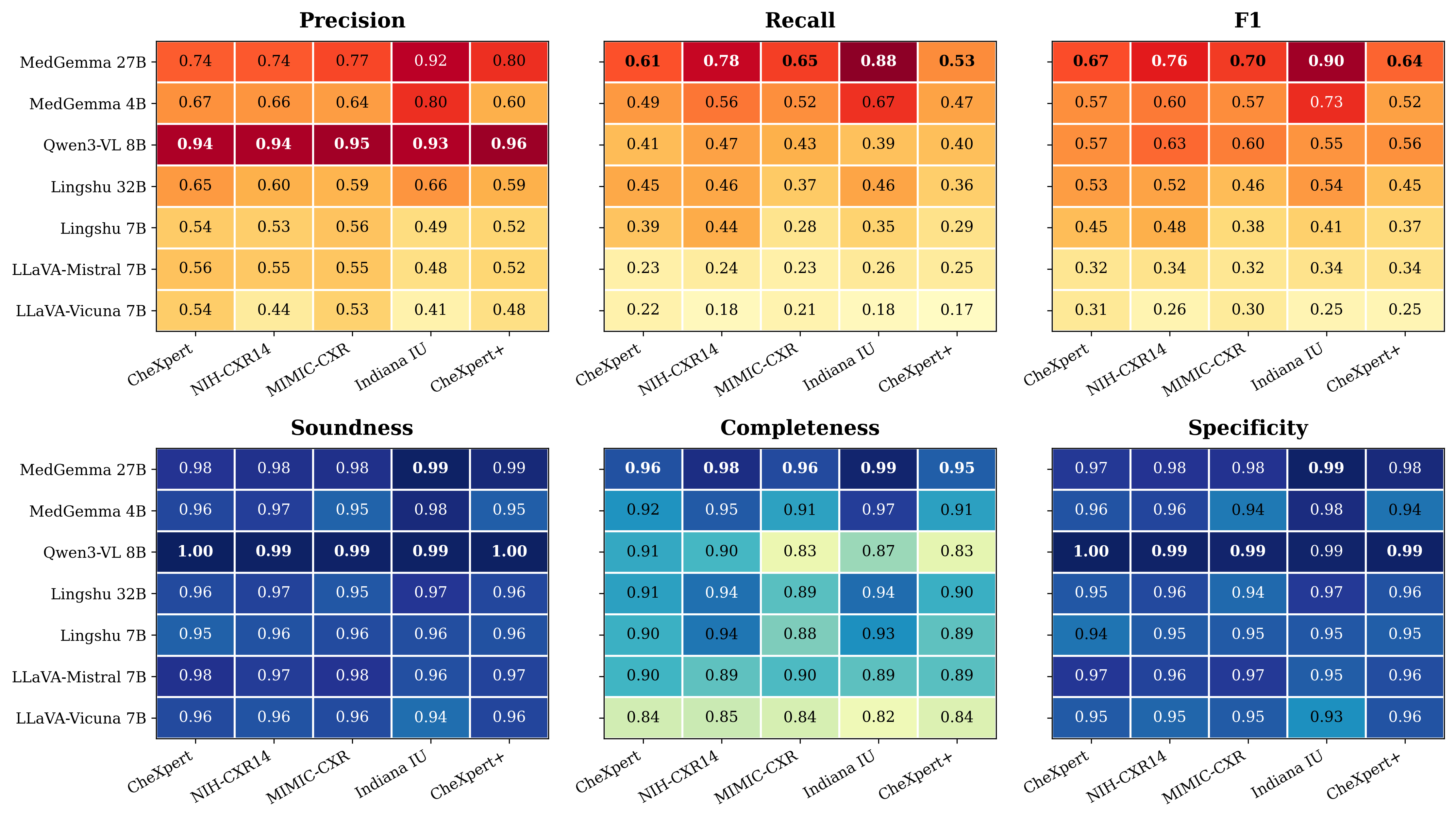}
\caption{Heatmap of model performance under neurosymbolic auditing across precision, recall, F1, soundness, completeness, and specificity, revealing distinct reasoning profiles across VLM families.}
    \label{fig:placeholder}
\end{figure}

\subsection{Impact of Symbolic Filtering (on Labeled Datasets)}
We evaluate the effect of solver-based verification on datasets with structured ground-truth labels by comparing diagnoses extracted from the raw VLM impression (\textsc{VLM}) with the subset of diagnoses that remain after symbolic entailment filtering (\textsc{Ours}). We define the verification delta as $\Delta = \textsc{Ours} - \textsc{VLM}$. 

Table~\ref{tab:vlm_z3_deltas} shows a consistent pattern across both CheXpert and NIH-CXR14: enforcing entailment increases \emph{soundness} for every model, while slightly reducing \emph{completeness}. This behavior is expected because \textsc{Ours} only retains diagnoses that are \emph{deductively licensed} by the evidence extracted from the Findings section under $\mathcal{K}$. As a result, some diagnoses that were previously stated in the impression, or that match ground-truth labels, may be removed if the necessary supporting evidence is not present (or not extracted) in $R_F$. Importantly, the decrease in completeness is uniformly small across models and datasets, indicating that verification does not broadly suppress diagnoses that are logically entailed; rather, it preserves the majority of entailed conclusions while improving consistency. Across all models, $\Delta$Precision is uniformly positive and $\Delta$Recall uniformly negative, indicating that symbolic verification acts as a conservative post-hoc filter that removes impression claims unsupported by the model's stated evidence, improving precision, internal consistency, and auditability at a small cost in sensitivity (recall).

\begin{table}[htbp]
\centering
\begin{adjustbox}{max width=\linewidth}
\begin{tabular}{
p{3.1cm}
S[table-format=1.4]
S[table-format=1.4]
S[table-format=+1.4]
S[table-format=1.4]
S[table-format=1.4]
S[table-format=+1.4]
S[table-format=+1.4]
S[table-format=+1.4]
}
\toprule
& \multicolumn{3}{c}{Soundness} & \multicolumn{3}{c}{Completeness} & \multicolumn{2}{c}{Other Deltas} \\
\cmidrule(lr){2-4}\cmidrule(lr){5-7}\cmidrule(lr){8-9}
Model
& {VLM} & {Ours} & {$\Delta$}
& {VLM} & {Ours} & {$\Delta$}
& {$\Delta$ Recall} & {$\Delta$ Prec.} \\
\midrule

\multicolumn{9}{l}{\textbf{CheXpert}} \\
\addlinespace[2pt]
\texttt{MedGemma-27B}       & 0.9301 & 0.9599 & +0.0298 & 0.8231 & 0.8107 & -0.0124 & -0.0601 & +0.0024 \\
\texttt{MedGemma-4B}        & 0.9115 & 0.9311 & +0.0196 & 0.8589 & 0.8465 & -0.0124 & -0.0600 & +0.0098 \\
\texttt{Llava-Mistral-7B}   & 0.9246 & 0.9684 & \best{+0.0438} & 0.8140 & 0.8062 & -0.0078 & -0.0379 & \best{+0.0736} \\
\texttt{Llava-Vicuna-7B}    & 0.9061 & 0.9458 & +0.0397 & 0.8212 & 0.8126 & -0.0086 & -0.0416 & +0.0327 \\
\texttt{Lingshu-7B}         & 0.9182 & 0.9494 & +0.0312 & 0.8436 & 0.8246 & -0.0190 & -0.0918 & +0.0034 \\
\texttt{Lingshu-32B}        & 0.9125 & 0.9426 & +0.0301 & 0.8379 & 0.8248 & -0.0131 & -0.0637 & +0.0161 \\
\texttt{Qwen3-VL-8B}        & 0.9631 & 0.9689 & +0.0058 & 0.8151 & 0.8128 & \best{-0.0023} & \best{-0.0111} & +0.0132 \\
\addlinespace[4pt]
\midrule

\multicolumn{9}{l}{\textbf{NIH-CXR14}} \\
\addlinespace[2pt]
\texttt{MedGemma-27B}       & 0.9505 & 0.9780 & +0.0275 & 0.9488 & 0.9460 & -0.0028 & -0.0334 & \best{+0.1784} \\
\texttt{MedGemma-4B}        & 0.9090 & 0.9343 & +0.0253 & 0.9462 & 0.9432 & -0.0030 & -0.0358 & +0.0433 \\
\texttt{Llava-Mistral-7B}   & 0.9223 & 0.9659 & \best{+0.0436} & 0.9318 & 0.9269 & -0.0049 & -0.0589 & +0.0708 \\
\texttt{Llava-Vicuna-7B}    & 0.8997 & 0.9385 & +0.0388 & 0.9335 & 0.9269 & -0.0066 & -0.0779 & +0.0018 \\
\texttt{Lingshu-7B}         & 0.9231 & 0.9585 & +0.0354 & 0.9403 & 0.9360 & -0.0043 & -0.0509 & +0.0844 \\
\texttt{Lingshu-32B}        & 0.9365 & 0.9665 & +0.0300 & 0.9395 & 0.9359 & -0.0036 & -0.0437 & +0.1015 \\
\texttt{Qwen3-VL-8B}        & 0.9582 & 0.9713 & +0.0131 & 0.9271 & 0.9263 & \best{-0.0008} & \best{-0.0104} & +0.0522 \\
\bottomrule
\end{tabular}
\end{adjustbox}
\vspace{1em}
\caption{Comparison of raw VLM impressions and our verified diagnoses on labeled datasets. Enforcing entailment increases soundness and precision while reducing completeness and recall.}
\label{tab:vlm_z3_deltas}
\end{table}


\section{Discussion}
\textbf{From Empirical Metrics to Deductive Validity.}
As we had mentioned earlier, autoregressive medical VLMs generate fluent reports, but maximum-likelihood decoding cannot guarantee deductive validity. Standard metrics (BLEU, ROUGE-L) penalize harmless paraphrasing, fail in reference-free deployment, and remain blind to explicit contradictions between findings and impressions. Framing consistency as formal satisfiability ($\Phi_V \wedge \mathcal{K} \models d$) allows deterministic mathematical verification, and reference-free runtime audits of clinical reasoning.

\textbf{Deductive Archetypes in Medical VLMs.}
Neurosymbolic auditing exposes three distinct behavioral archetypes across architectures:
\begin{itemize}
    \item Conservative Observers (e.g., Qwen3-VL-8B): Attain near-perfect soundness ($S \ge 0.99$) but constrained completeness ($C \approx 0.87$), rarely hallucinating but frequently omitting diagnoses logically forced by their own findings.
    \item Stochastic Hallucinators (e.g., LLaVA-Vicuna-7B): Exhibit degraded soundness ($S \approx 0.95$) and precision, generating impression claims that contradict or lack support in the findings, elevating automation bias risks.
    \item Balanced Models (e.g., MedGemma-27B): Achieve balanced scores ($S = 0.981, C = 0.972$), showing that domain-specific scaling improves deductive coherence.
\end{itemize}
Operationally, conservative models merely require verbalization prompting, whereas stochastic models necessitate active safeguard interventions.

\textbf{Post-Hoc Deductive Safeguards.}
Enforcing solver-backed entailment filtering uniformly increases precision and soundness across CheXpert and NIH-CXR14. The accompanying minor reduction in completeness and recall is clinically advantageous: eliminating ungrounded diagnostic claims suppresses false positives, preventing unwarranted invasive procedures, redundant follow-up imaging, and premature diagnostic closure.

\textbf{Assume-Guarantee Verification Contract.}
Our framework operates under a formal assume-guarantee contract: conditional on accurate visual perception and faithful text autoformalization $\mathcal{T}$, the SMT solver guarantees that the impression provably entails from the findings under clinical knowledge base $\mathcal{K}$. This establishes an interpretable division of labor where radiologists verify perceptual assertions while the solver guarantees deductive validity. Additionally, the Closed-World Assumption naturally mirrors clinical documentation standards where unmentioned findings denote absent pathology.

\textbf{Feasibility and Future Horizons.}
Because propositional encodings yield compact constraint sets, Z3 resolves queries in milliseconds, enabling real-time integration into hospital PACS workflows, given fast inference. Future extensions include utilizing LEAN encoded general purpose electronic health record integration with higher reusability and custom-tactics with meta-programming to guarantee deductive soundness of LLM based clinical decision support systems.
\section{Conclusion}
\label{sec:conclusion}
We present a neurosymbolic verification framework for radiology VLM reports that audits whether Impression diagnoses are logically supported by the model’s own Findings using autoformalization, a clinician-audited knowledge base, and SMT solving. Our results show that this enables reference-free measurement of deductive reliability, exposes clinically important reasoning errors missed by lexical metrics, and provides a practical post-hoc safeguard that consistently improves soundness and precision on labeled datasets with only modest recall and completeness trade-offs. More broadly, this shifts evaluation from surface text similarity toward verifiable internal consistency, offering a concrete path to safer and more auditable generative clinical assistants under explicit assume-guarantee conditions common in the domain of formal verification.

\begin{ack}
This work was supported in part by NSF research grants \#2112606 and \#2117439.
\end{ack}

\bibliographystyle{plainnat}
\bibliography{references}

\newpage
\section*{NeurIPS Paper Checklist}

The checklist is designed to encourage best practices for responsible machine learning research, addressing issues of reproducibility, transparency, research ethics, and societal impact. Do not remove the checklist: {\bf The papers not including the checklist will be desk rejected.} The checklist should follow the references and follow the (optional) supplemental material.  The checklist does NOT count towards the page
limit. 

Please read the checklist guidelines carefully for information on how to answer these questions. For each question in the checklist:
\begin{itemize}
    \item You should answer \answerYes{}, \answerNo{}, or \answerNA{}.
    \item \answerNA{} means either that the question is Not Applicable for that particular paper or the relevant information is Not Available.
    \item Please provide a short (1--2 sentence) justification right after your answer (even for \answerNA). 
\end{itemize}

{\bf The checklist answers are an integral part of your paper submission.} They are visible to the reviewers, area chairs, senior area chairs, and ethics reviewers. You will also be asked to include it (after eventual revisions) with the final version of your paper, and its final version will be published with the paper.

The reviewers of your paper will be asked to use the checklist as one of the factors in their evaluation. While \answerYes{} is generally preferable to \answerNo{}, it is perfectly acceptable to answer \answerNo{} provided a proper justification is given (e.g., error bars are not reported because it would be too computationally expensive'' or ``we were unable to find the license for the dataset we used''). In general, answering \answerNo{} or \answerNA{} is not grounds for rejection. While the questions are phrased in a binary way, we acknowledge that the true answer is often more nuanced, so please just use your best judgment and write a justification to elaborate. All supporting evidence can appear either in the main paper or the supplemental material, provided in appendix. If you answer \answerYes{} to a question, in the justification please point to the section(s) where related material for the question can be found.

IMPORTANT, please:
\begin{itemize}
    \item {\bf Delete this instruction block, but keep the section heading ``NeurIPS Paper Checklist"},
    \item  {\bf Keep the checklist subsection headings, questions/answers and guidelines below.}
    \item {\bf Do not modify the questions and only use the provided macros for your answers}.
\end{itemize}


\begin{enumerate}

\item {\bf Claims}
    \item[] Question: Do the main claims made in the abstract and introduction accurately reflect the paper's contributions and scope?
    \item[] Answer: \answerTODO{} 
    \item[] Justification: \justificationTODO{}

\item {\bf Limitations}
    \item[] Question: Does the paper discuss the limitations of the work performed by the authors?
    \item[] Answer: \answerTODO{} 
    \item[] Justification: \justificationTODO{}

\item {\bf Theory assumptions and proofs}
    \item[] Question: For each theoretical result, does the paper provide the full set of assumptions and a complete (and correct) proof?
    \item[] Answer: \answerTODO{} 
    \item[] Justification: \justificationTODO{}

    \item {\bf Experimental result reproducibility}
    \item[] Question: Does the paper fully disclose all the information needed to reproduce the main experimental results of the paper to the extent that it affects the main claims and/or conclusions of the paper (regardless of whether the code and data are provided or not)?
    \item[] Answer: \answerTODO{} 
    \item[] Justification: \justificationTODO{}

\item {\bf Open access to data and code}
    \item[] Question: Does the paper provide open access to the data and code, with sufficient instructions to faithfully reproduce the main experimental results, as described in supplemental material?
    \item[] Answer: \answerTODO{} 
    \item[] Justification: \justificationTODO{}

\item {\bf Experimental setting/details}
    \item[] Question: Does the paper specify all the training and test details (e.g., data splits, hyperparameters, how they were chosen, type of optimizer) necessary to understand the results?
    \item[] Answer: \answerTODO{} 
    \item[] Justification: \justificationTODO{}

\item {\bf Experiment statistical significance}
    \item[] Question: Does the paper report error bars suitably and correctly defined or other appropriate information about the statistical significance of the experiments?
    \item[] Answer: \answerTODO{} 
    \item[] Justification: \justificationTODO{}

\item {\bf Experiments compute resources}
    \item[] Question: For each experiment, does the paper provide sufficient information on the computer resources (type of compute workers, memory, time of execution) needed to reproduce the experiments?
    \item[] Answer: \answerTODO{} 
    \item[] Justification: \justificationTODO{}
    
\item {\bf Code of ethics}
    \item[] Question: Does the research conducted in the paper conform, in every respect, with the NeurIPS Code of Ethics \url{https://neurips.cc/public/EthicsGuidelines}?
    \item[] Answer: \answerTODO{} 
    \item[] Justification: \justificationTODO{}

\item {\bf Broader impacts}
    \item[] Question: Does the paper discuss both potential positive societal impacts and negative societal impacts of the work performed?
    \item[] Answer: \answerTODO{} 
    \item[] Justification: \justificationTODO{}
    
\item {\bf Safeguards}
    \item[] Question: Does the paper describe safeguards that have been put in place for responsible release of data or models that have a high risk for misuse (e.g., pre-trained language models, image generators, or scraped datasets)?
    \item[] Answer: \answerTODO{} 
    \item[] Justification: \justificationTODO{}

\item {\bf Licenses for existing assets}
    \item[] Question: Are the creators or original owners of assets (e.g., code, data, models), used in the paper, properly credited and are the license and terms of use explicitly mentioned and properly respected?
    \item[] Answer: \answerTODO{} 
    \item[] Justification: \justificationTODO{}

\item {\bf New assets}
    \item[] Question: Are new assets introduced in the paper well documented and is the documentation provided alongside the assets?
    \item[] Answer: \answerTODO{} 
    \item[] Justification: \justificationTODO{}

\item {\bf Crowdsourcing and research with human subjects}
    \item[] Question: For crowdsourcing experiments and research with human subjects, does the paper include the full text of instructions given to participants and screenshots, if applicable, as well as details about compensation (if any)? 
    \item[] Answer: \answerTODO{} 
    \item[] Justification: \justificationTODO{}

\item {\bf Institutional review board (IRB) approvals or equivalent for research with human subjects}
    \item[] Question: Does the paper describe potential risks incurred by study participants, whether such risks were disclosed to the subjects, and whether Institutional Review Board (IRB) approvals (or an equivalent approval/review based on the requirements of your country or institution) were obtained?
    \item[] Answer: \answerTODO{} 
    \item[] Justification: \justificationTODO{}

\item {\bf Declaration of LLM usage}
    \item[] Question: Does the paper describe the usage of LLMs if it is an important, original, or non-standard component of the core methods in this research? Note that if the LLM is used only for writing, editing, or formatting purposes and does \emph{not} impact the core methodology, scientific rigor, or originality of the research, declaration is not required.
    \item[] Answer: \answerTODO{} 
    \item[] Justification: \justificationTODO{}

\end{enumerate}

\end{document}